\def\year{2022}\relax
\documentclass[letterpaper]{article} 
\usepackage{aaai22}  
\usepackage{times}  
\usepackage{helvet}  
\usepackage{courier}  
\usepackage[hyphens]{url}  
\usepackage{graphicx} 
\usepackage{natbib}  
\usepackage{caption} 
\DeclareCaptionStyle{ruled}{labelfont=normalfont,labelsep=colon,strut=off} 
\usepackage{algorithm}
\usepackage{algorithmic}

\usepackage{times}
\usepackage{epsfig}
\usepackage{graphicx}
\usepackage{amsmath}
\usepackage{amssymb}
\usepackage{bm}
\usepackage{subfigure} 
\usepackage{xr}
\usepackage{amsthm}
\newtheorem{prop}{Proposition}
\usepackage{algorithm,algorithmic}

\usepackage[pagebackref=true,breaklinks=true,letterpaper=true,colorlinks,bookmarks=false]{hyperref}

\usepackage{newfloat}
\usepackage{listings}
\floatstyle{ruled}
\newfloat{listing}{tb}{lst}{}
\floatname{listing}{Listing}
\title{Exploiting Invariance in Training Deep Neural Networks}

\author {Chengxi Ye, Xiong Zhou, Tristan McKinney, Yanfeng Liu, Qinggang Zhou,Fedor Zhdanov
}
\affiliations {
Amazon Web Services\\
chengxye,xiongzho,tristamc,liuyanfe,qingganz@amazon.com, fedor.zhdanov@rhul.ac.uk
}

\begin{document}

\maketitle


\begin{abstract}
Inspired by two basic mechanisms in animal visual systems, we introduce a feature transform technique that imposes invariance properties in the training of deep neural networks. The resulting algorithm requires less parameter tuning, trains well with an initial learning rate $1.0$, and easily generalizes to different tasks. We enforce scale invariance with local statistics in the data to align similar samples at diverse scales. To accelerate convergence, we enforce a $GL(n)$-invariance property with global statistics extracted from a batch such that the gradient descent solution should remain invariant under basis change. Profiling analysis shows our proposed modifications takes $\sim 5\%$ of the computations of the underlying convolution layer. Tested on convolutional networks and transformer networks, our proposed technique requires fewer iterations to train, surpasses all baselines by a large margin, seamlessly works on both small and large batch size training, and applies to different computer vision and language tasks.
\end{abstract}

\section{Introduction}

The pupillary light reflex constricts the pupil in bright light and dilates the pupil in dim light~\cite{bear2020neuroscience}. This mechanism controls the amount of light passing into the eye, allowing common features to be extracted from signals of different scales. In addition, retinal receptive fields use center-surround structures to filter and sharpen images~\cite{hubel1962receptive} (Supp. Fig.~\ref{fig:hubel-wiesel} a, b)
, removing a bell-shaped autocorrelation present in real-world visual signals. Intriguingly, all other receptive field configurations discovered in Hubel and Wiesel's seminal research (Supp. Fig.~\ref{fig:hubel-wiesel} c, d, e) have found artificial analogs in the first layer filters learned by modern convolutional neural networks~\cite{zeiler2014visualizing} (Fig. 5 a,b).

\begin{figure}[hbt!]
\subfigure[]{\includegraphics[width=.48\columnwidth]{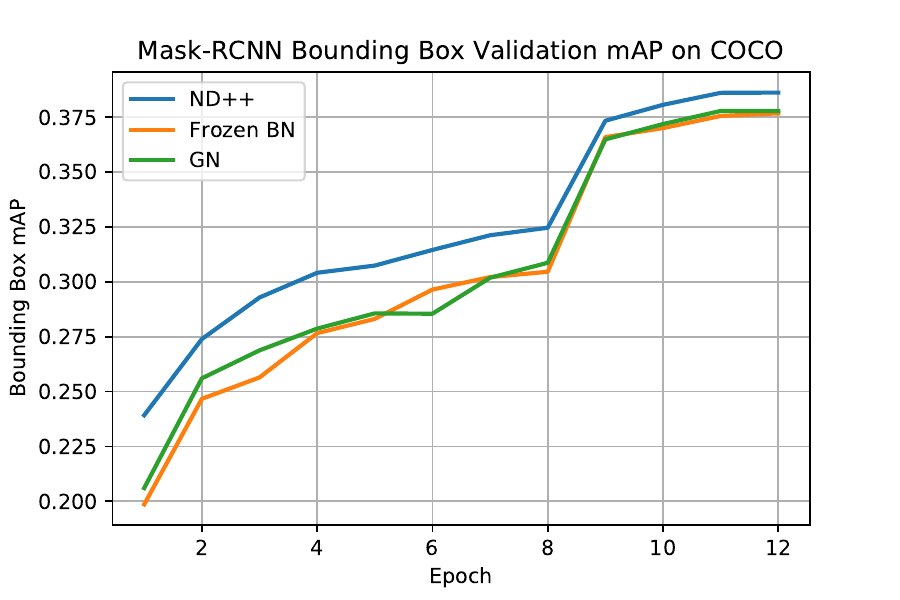}}
\subfigure[]{\includegraphics[width=.48\columnwidth]{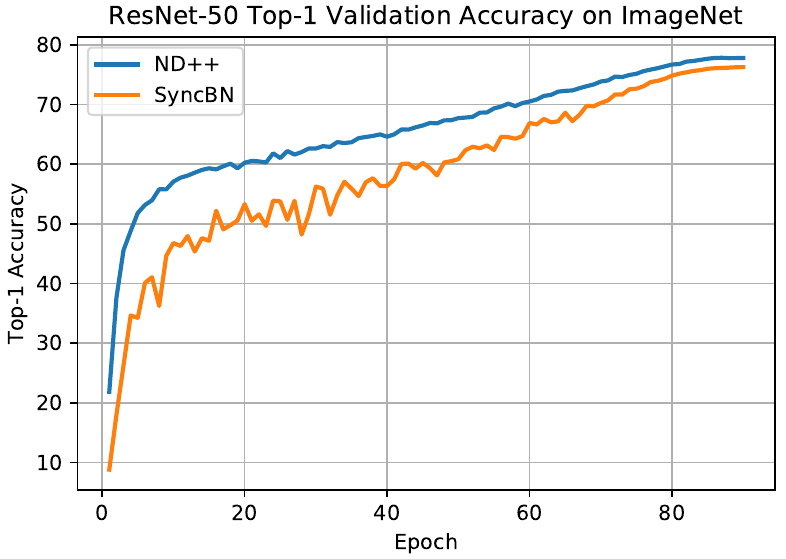}}
\subfigure[]{\includegraphics[width=.48\columnwidth]{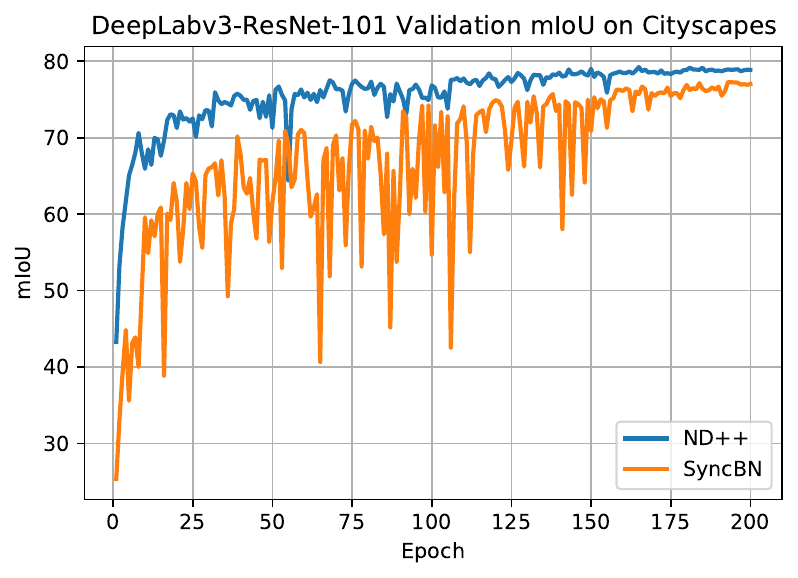}}
\subfigure[]{\includegraphics[width=.48\columnwidth]{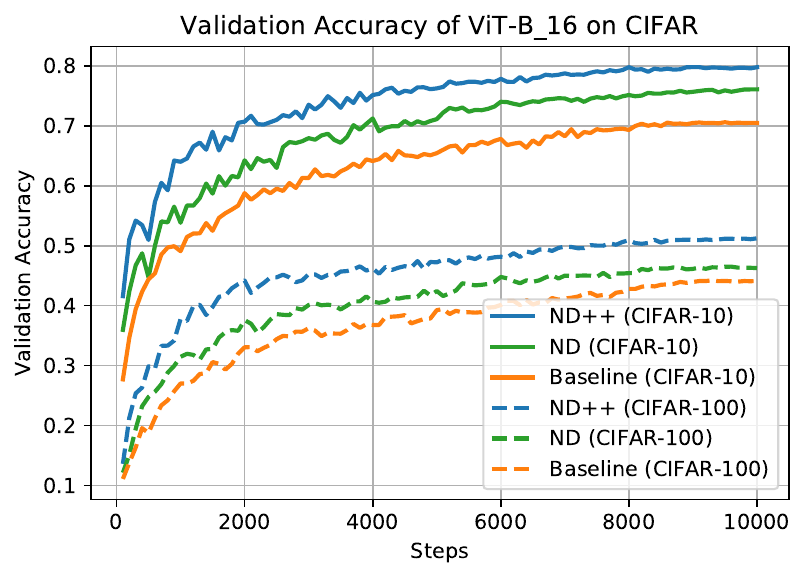}}
\caption[front page]{(a) Validation AP curves of a Mask R-CNN network when trained with network deconvolution++ (ND++), frozen batch normalization and group normalization. (b) Top-1 validation accuracy of ResNet-50 when trained on ImageNet with ND++ and SyncBN using batch size of $2048$, the training finishes better and faster than using batch size $256$ using the same hardware. (c) mIoU curves of DeepLabv3 with a ResNet-101 backbone trained on the Cityscapes dataset using ND++ and SyncBN. ND++ consistently outperforms baselines and produces more stable mIoU curves. (d) ND++ significantly improves upon ND when training ViT-B\_16 on the CIFAR datasets.}
\label{fig:frontpage}
\end{figure}

While convolutional networks continue to push the envelope in computer vision tasks, state-of-the-art training recipes are still limited by scope and scale. Specifically, when moving from image classification to object detection, different normalization techniques need to be used. Most algorithms perform well at a specific scale, and there is usually a significant drop in accuracy when the training batch size is too large or too small~\cite{GroupNorm,singh2020filter}.

In this paper, we conduct a study of a one-layer linear network to understand the origin of these limitations. Drawing inspiration from this study, the structure of animal visual systems mentioned above, and recent related work~\cite{Ye2020Network}, we derive two invariance properties that enhance the training of deep neural networks. We implement cross-GPU synchronization to aggregate the computation required to enforce the invariance, surpassing the widely-used synchronized batch normalization~\cite{DBLP:journals/corr/abs-1711-07240} method significantly. This implementation supports both small batch and large batch training without algorithm change. By enforcing the invariance properties at every layer of the network, we accelerate training convergence and surpass baseline accuracy by a large margin on the ImageNet~\cite{deng2009imagenet}, MS COCO~\cite{lin2014microsoft}, and Cityscapes~\cite{cordts2016cityscapes} datasets for image classification, object detection, and semantic segmentation, respectively (Fig.~\ref{fig:frontpage}). In our supplementary materials, we also show promising results for training transformers on multiple vision and language tasks.

Our main findings are the following:
\begin{itemize}
\item We propose a drop-in modification \textit{before} the linear layers in a network to explicitly enforce these two invariance properties, significantly reducing training iterations and surpassing baseline accuracy by a large margin.
\item Training with this modification is robust to different optimizer configurations; using the theoretically optimal learning rate for the linear case of $1.0$ can successfully train a wide range of deep networks.
\item Our implementation takes $\sim 5\%$ (Supp. Table~\ref{tab:runtime}) of the cost of the underlying convolution layer and incorporates a cross-GPU synchronization which surpasses synchronized batch normalization by a large margin in both small and large batch regime.
\item The benefits of the proposed modification generalize to emerging architectures in vision and language.
\end{itemize}

\section{Background}

\subsection{Backward Correction Methods}

The complicated loss landscapes~\cite{DBLP:journals/corr/abs-1712-09913} of neural networks create numerous challenges for training. Deep neural networks are generally over-parameterized. The presence of strong correlation between features induces areas of pathological curvature in the landscape and inhibits effective training. Small gradients are common in these pathological regions, and the problem is exacerbated by small linear layers and common activation functions. These issues were traditionally addressed by correcting the gradients~\cite{nocedal2006numerical,martens2010deep,DBLP:journals/corr/abs-1708-00631}, that is, by modifying the backward pass. The most popular methods normalize the gradient scale to avoid the vanishing gradient problem and smooth the direction by using previous gradients as a momentum term~\cite{kingma2014adam}. More advanced methods, such as Newton's method, use approximate curvature information to modify the gradient direction (Fig.~\ref{fig:correction}). However, high computational costs limit these algorithms to small-scale problems~\cite{DBLP:journals/corr/MartensG15,NIPS2015_5953}. They have not been widely shown to surpass first-order gradient descent methods.

\subsection{Forward Correction Methods}

\begin{figure}
\subfigure[]{\includegraphics[width=.3\columnwidth]{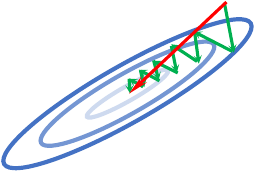}}
\subfigure[]{\includegraphics[width=.33\columnwidth]{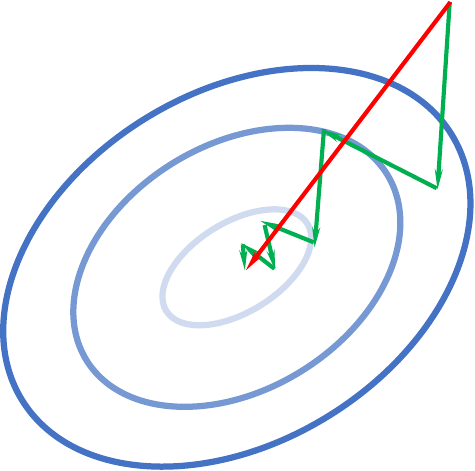}}
\subfigure[]{\includegraphics[width=.33\columnwidth]{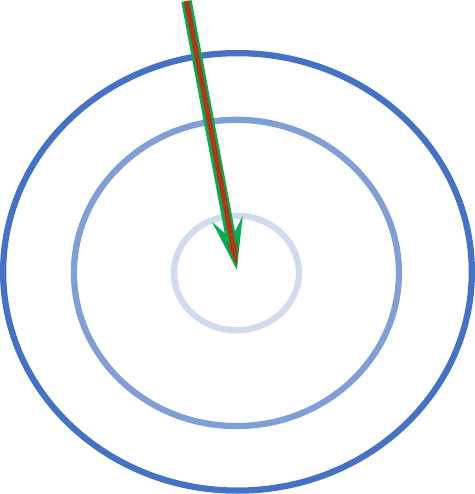}}
\caption[correction methods]{A toy picture of loss landscapes and gradient steps. The red arrows correspond to the gradients found through backward correction or Newton's method and the green arrows correspond to the gradients used in gradient descent iterations. (a) Highly correlated, non-standardized data. (b) Correlated but standardized data. (c) With standardized and uncorrelated data, vanilla gradient descent coincides with Newton's method (Proposition ~\ref{prop1}). }\label{fig:correction}
\end{figure}

Forward transforms provide an alternative approach to address the challenges in training deep neural networks~\cite{huang2020normalization}. Batch normalization~\cite{ioffe2015batch} is a common and powerful example that standardizes the distribution of features in each dimension. This stretching with a diagonal matrix works perfectly only for uncorrelated, axis-aligned features, suggesting that the optimality of standardization \textit{depends on the choice of basis}. Given the inherent correlation in real-world data, the feature covariance matrix is generally ill-conditioned and not diagonal. As a result, gradient descent training usually takes unnecessary steps towards the solution (Fig.~\ref{fig:correction}(b)). Recently, more accurate transforms have been proposed to utilize the covariance matrix and remove the pairwise correlation between features~\cite{huang2018decorrelated,DBLP:journals/corr/abs-1809-08625,huang2019iterative,pan2019switchable,Ye2020Network}. If the features are transformed to be axis-aligned and have unit-variance, the loss landscape is more isotropic. Gradient descent converges more quickly and accurately (Fig.~\ref{fig:correction}(c)). 

\section{Motivations}

We start our investigation with a toy example of a one-layer network (Eq.~\ref{MSE_loss}). Rigorous analysis of this simple setup reveals a surprising depth of insight into the fundamental training problem. The analysis explains the utility of various modern techniques to facilitate training and leads us to a more advantageous design.

Assume we are given a linear regression problem with a mean squared error loss (Eq.~\ref{MSE_loss}). $\hat{y}$ is the continuous or discrete response data to be regressed. This formulation can be used for prediction or classification. In the typical setting, the output $y=X w$ is given by multiplying the inputs $X$ with an unknown weight vector $w$ for which we are solving. Let $N$ be the number of samples, $d$ the feature dimension, and $X$  either the $N \times d$ or augmented $N \times (d+1) $ data matrix $(X|\bm{1})$, if we include a bias.

\begin{equation}
Loss_{MSE}=\frac{1}{2}E(| y -\hat{y} |^2)=\frac{1}{2N}\| Xw -\hat{y} \|^2.
\label{MSE_loss}
\end{equation}

\subsection{Local Statistics vs Global Statistics}

Given a $N \times d$ data matrix $X$, we refer to the column statistics as the \textit{global statistics} and the row statistics as \textit{local statistics}. Mini-batch statistics represent an approximation to the global statistics of the whole dataset. Batch normalization~\cite{ioffe2015batch} standardizes the $d$ column vectors. This can be visualized as a coordinate transform that stretches the data along each axis based on global statistics (Fig.~\ref{fig:correction} (a) to (b)). Training then solves for \textit{a new set of weights $w_{BN}$} in this transformed space.

On the other hand, sample-based normalization~\cite{ba2016layer,GroupNorm,singh2020filter} stretches each sample by removing the scale and bias in each row of $X$ according to the local statistics. Viewed from the original space, training corresponds to finding \textit{$N$ sets of sample-variant weights $\{w_i\}$} that both stretch the samples and fit the model. 

\subsection{$GL(n)$-invariance}

One may ask whether the solution of Eq.~\ref{MSE_loss} found using a given algorithm changes under a change of basis. That is, if we use any invertible linear transform, i.e., a member of the general linear group $GL(n)$~\cite{artin2011algebra}, to transform the features, will we simply reach an equivalent solution in a different coordinate system? If the solution is invariant under the operation of $GL(n)$, we will call the training algorithm $GL(n)$-invariant.

Assuming $X^{t}X$ is invertible for our toy problem, the unique solution can be found by setting the gradient to $0$, $\frac{\partial Loss}{\partial w}=\frac{1}{N}X^t(Xw-\hat{y})=0$,
\begin{equation}
    w={(X^t X)}^{-1} X^{t} \hat{y}.
\label{normal_eq}
\end{equation}

Let us consider the impact of different correction methods on the ability of gradient descent-based algorithms to find this solution.

\textbf{Case 1 (backward correction/Newton's method)}: For simplicity, suppose we start from $w_0=0$, then $\frac{\partial Loss}{\partial w}|_{w_0}= -\frac{1}{N}X^t\hat{y}$. Letting $H=\frac{1}{N}X^t X=\nabla^2 w$, we see from Eq.~\ref{normal_eq} that $w=-H^{-1} \frac{\partial Loss}{\partial w}$. This derivation shows us how the gradients can be manipulated in the backward fashion to accelerate convergence~\cite{DBLP:journals/corr/MartensG15,NIPS2015_5953,DBLP:journals/corr/abs-1708-00631}: (1) approximate the curvature with $H$ and apply an inverse correction to \textit{decorrelate the gradient}, $H^{-1}\frac{\partial Loss}{\partial w}$, then (2) take a descent step using a learning rate of $1.0$: $w=w_0-1.0\cdot H^{-1}\frac{\partial Loss}{\partial w}$. $GL(n)$ invariance can be achieved because the optimal solution can be found in one step by following the corrected negative gradient in \textit{any basis} (found as the red arrows in Fig.~\ref{fig:correction}).

\textbf{Case 2 (forward correction)}: Instead of the Newton's method, we adopt a forward correction point of view. Simple forward corrections include standardization using global statistics, which results in standardized columns, while using local statistics results in standardized rows. In terms of convergence rate, using global statistics is slightly superior, as after correction $\frac{1}{N}H=\frac{1}{N}X^t{X}$ is guaranteed to have unit diagonal. Therefore, the loss landscape has better statistical properties and convergence is accelerated.\footnote{Normalization with local statistics works on $XX^t$, which is less related to the convergence.} However, standardization with either local or global statistics does not remove correlations between features. On an elongated energy landscape, gradient descent algorithms generally do not converge in one step and the optimal solution cannot be found under an arbitrary basis change. The following proposition demonstrates that forward correction can achieve the power of Newton's method if we (1) \textit{decorrelate the feature columns with global statistics} and (2) use a learning rate $1.0$.

\begin{prop}
\label{prop1}
The optimal solution can be found in one step with the forward correction if and only if the feature columns are standardized and uncorrelated. 
\end{prop}

\begin{proof}
With these features we have $\frac{1}{N}X^tX=I$, and the optimal explicit solution (Eq.~\ref{normal_eq}) simplifies to $w=\frac{1}{N}X^t\hat{y}$. After one iteration of gradient descent, $w_{new}=w_{old}-\eta \frac{1}{N}(X^t{X}w_{old}-X^t \hat{y})$. By substituting $\frac{1}{N}X^t{X} = I$, we find that $\eta=1.0$ is optimal and yields convergence in a single iteration. 
\end{proof}

\subsection{Scale Invariance}
In our toy problem, the output scales linearly with the input, $(aX)w=a(Xw)$. In fact, this property generalizes to common networks with linear layers and ReLU-like activation functions. 
If we consider a task like bounding box prediction, this suggests that we can scale the box size by simply changing the input brightness, which should clearly be avoided. Through the pupillary light reflex, animal visual systems introduce a scale invariance property so that differently scaled inputs generate similarly scaled features. Although transforming using global statistics can guarantee $GL(n)$-invariance as discussed above, it does not result in scale invariance. Since scale invariance is a concern in some tasks, we utilize both global and local statistics (where samples are transformed individually) in the algorithm. We defer the explanation of details to the next section. 

\subsection{Optimizing the Formulation}
Influenced by the batch normalization, most normalization algorithms~\cite{ba2016layer,GroupNorm,singh2020filter,huang2018decorrelated, huang2019iterative,pan2019switchable} adopt a post-normalization design, normalizing $y=Xw$ after the linear transform rather than normalizing $X$. Since post-normalization restricts the representation power of $y$, two extra parameters are introduced to remove this limitation: $Loss_{MSE}=\frac{1}{2N}||(Xw-\mu)\Sigma^{-1}\gamma +\beta-\hat{y}||^2$. As a result, the toy problem contains more than necessary parameters ($w,\gamma,\beta$ instead of $w$) to be optimized. 
However, the optimal convergence property that we find on the toy problem no longer holds with the introduction of the redundancy. This motivates us to optimize our design to remove the redundancy while maintaining full representation power.

\section{Enforcing Invariance in Training Neural Networks}
\label{sec:invariance}

We demonstrate how the problems in the previous section can be naturally resolved. Some complementary information can be found in recent works~\cite{Ye2020Network}.

To improve the network training, we enhance the linear transform in neural networks by applying the principles derived from the toy problem. We discuss how to enhance three common linear transform layers, starting from analyzing the data matrix $X$ in the network.


Case 1. The data matrix $X$ of a fully-connected layer is constructed straightforwardly by stacking $N$ rows of $d$ or $d+1$ dimensional feature vectors. We put our emphasis on the data matrix construction of convolution and (cross-)correlation layers.

Case 2. A convolution operation can be expressed in multiple ways in the spatial domain. We can expand the kernel of size $k$ into a $(n-k+1) \times n$ convolution matrix $W$, or unroll the overlapping windows of $x$ into a $(n-k+1) \times k$ data matrix $X$: $y=x\ast w=Wx =Xw_{flipped}$. We provide a $1d$ convolution example using $k=3$ (in the `valid' mode according to Matlab terminology):
\begin{gather*}
\begin{pmatrix}
y_1\\
y_2\\
...\\
y_{n-k}\\
y_{n-k+1} \\
\end{pmatrix}
=
\begin{pmatrix}
    w_3&w_2&w_1 \\
    &w_3&w_2&w_1\\
     &&\ddots\\
     &&&w_3&w_2&w_1\\
  \end{pmatrix}
\begin{pmatrix}
x_1\\
x_2\\
...\\
x_{n-1}\\
x_{n} \\
\end{pmatrix}
\\=
\begin{pmatrix}
x_1 & x_2 & x_3\\
x_2 & x_3 & x_4\\
x_3 & x_4 & x_5\\
...\\
x_{n-3} & x_{n-2} & x_{n-1}\\
x_{n-2} & x_{n-1} & x_{n}\\
\end{pmatrix}
\begin{pmatrix}
w_3\\
w_2\\
w_1 \\
\end{pmatrix}
\end{gather*}

Case 3. The correlation operation (also called transposed convolution, or sometimes misnamed as deconvolution) is the adjoint operation of the convolution, and involves padding the data by $k-1$ on each side (corresponding to the `full' mode in Matlab terminology) and using the unflipped kernel. The data matrix of correlation only differs from that in the convolution by the zero padding and is omitted here.
\begin{gather*}
\begin{pmatrix}
y_1\\
y_2\\
...\\
y_{n-1}\\
y_{n} \\
\end{pmatrix}
=
\begin{pmatrix}
     w_1&w_2&w_3\\
     &w_1&w_2&w_3\\
     &&\ddots\\
     &&&w_1&w_2&w_3\\
  \end{pmatrix}
\begin{pmatrix}
0\\
0\\
x_1\\
x_2\\
...\\
x_{n-k+1} \\
0\\
0\\
\end{pmatrix}
\end{gather*}

By closely looking at the different columns in the convolution/correlation data matrix $X$ (in case 2), one can find a rarely discussed problem in the training of networks. Since real-world data exhibits strong autocorrelation, and the neighboring columns of the data matrix correspond to shifting the signal by one pixel, the feature dimensions in the convolution/correlation data matrices are \textit{heavily correlated}. Therefore gradient descent training cannot converge efficiently with existing standardization techniques (Fig.~\ref{fig:correction} (b)) or just by decorrelating the feature channels in a layer.

\subsection{Connection with Frequency Domain Normalization}
\begin{figure}[hbt!]
\centering
\subfigure[]{\includegraphics[width=.2\columnwidth]{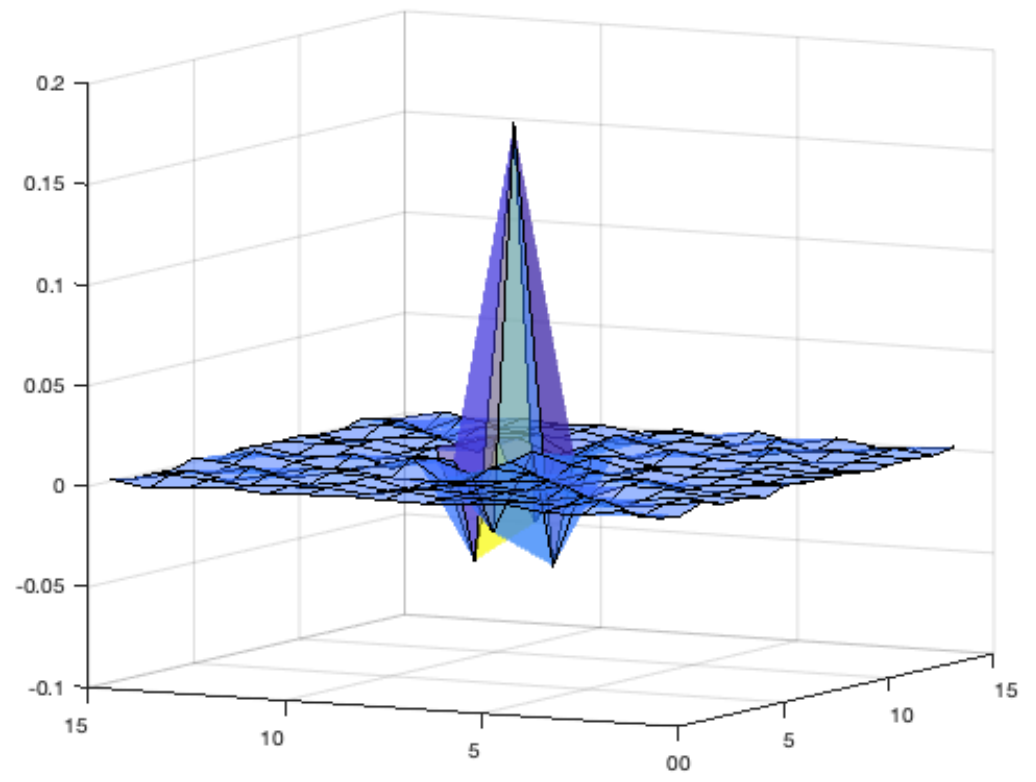}}
\subfigure[]{\includegraphics[width=.7\columnwidth]{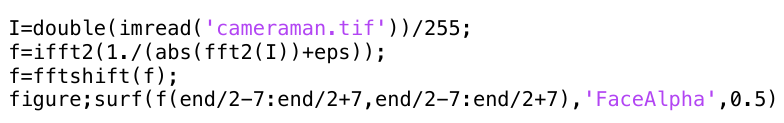}}
\caption[center surround]{Visualizing a frequency domain deconvolution filter in the spatial domain leads to a structure that has positive center values and negative surround values, resembling an on-center cell in animal visual systems~\cite{hubel1962receptive} (Supp. Fig.~\ref{fig:hubel-wiesel}(a)).}
\label{fig:center_surround}
\end{figure}

In signal processing, removing such pixel-wise autocorrelation has a long history and can be achieved through a process called whitening deconvolution~\cite{10.5555/1076432}. This correlation removal process can be generalized to every layer of a convolution neural network and we call this procedure network deconvolution. Usually this is achieved in the frequency domain through spectral whitening, i.e. a normalization in the frequency domain: $\frac{\mathcal{F}(x)}{|\mathcal{F}(x)|}=\frac{\mathcal{F}(x)}{\sqrt{\mathcal{F}^{H}(x)\cdot \mathcal{F}(x)}}$. This elegant normalization shows a profound insight: the optimal standardization for the convolution operation should in fact be carried out in the frequency domain. Note that frequency multiplication corresponds to a spatial domain convolution, and the frequency division corresponds to a spatial domain \textit{deconvolution}. 
Moreover, the deconvolution kernel $\mathcal{F}^{-1}(\frac{1}{|\mathcal{F}(x)|})$ has been found to resemble the center-surround structures in animal visual systems (Supp. Fig.~\ref{fig:hubel-wiesel}(a,b))~\cite{hubel1962receptive,Hyvrinen:2009:NIS:1572513,Ye2020Network}. Four lines of Matlab code are provided here for a quick verification (Fig.~\ref{fig:center_surround}).

\subsection{Spatial Domain Implementation}

\begin{algorithm}[]
\caption[ND++]{Proposed Computation for a Convolution Layer}
\begin{algorithmic}[1]
    \STATE{\textbf{Input:} Feature tensor $X_{N\times C\times H\times W}$ }
    \STATE{Scale features along first dimension $X_{scaled}=SX$}
    \STATE{Convert feature channels to data matrices along second dimension}
    \FOR{$i \in \{1,...,C\}$}
      \STATE{$X_i=im2col(X_{scaled}[:,i,:,:])$}
    \ENDFOR
    \STATE{$X=[X_1,...,X_C]$ \%Horizontally Concatenate}
    \STATE{$Cov=\frac{1}{N} X^t X$} \%Cross GPU Sync
    \STATE{$D\approx (Cov+\epsilon\cdot I)^{-\frac{1}{2}}$ \%via Inverse Newton Iterations }
    \STATE{\textbf{Output:} $Y = Conv(SX,DW)$ }

\end{algorithmic}
\label{ND++}
\end{algorithm}

For a practical implementation for convolutional networks, it is better to carry out the deconvolution computation in the spatial domain using the data matrix $X$ to match the small kernels that are used in practice. Specifically we calculate the covariance matrix $Cov=\frac{1}{N}X^tX$, find its unique principal inverse square root $D=Cov^{-\frac{1}{2}}$, and then use this matrix to transform the data matrix: $X_{0}=X\cdot D$. Note that $\frac{1}{N}X_{0}^tX_{0}$=$Cov^{-\frac{1}{2}}\cdot Cov\cdot Cov^{-\frac{1}{2}}=I$. As the transformed features are uncorrelated and standardized, faster convergence is achieved in this corrected space (Fig.~\ref{fig:correction}(c)). 
We track the correction $D$ during training and use the running average during testing.

For a convolution/correlation layer with $C$ channels, we construct the data matrix for each channel then horizontally concatenate the $C$ data matrices into a wider matrix (with $Ck$/$Ck^2$ columns for $1d$/$2d$). Deconvolution with this wider matrix removes the correlation of $k$/$k^2$ \textit{nearby pixels} and $C$ \textit{feature channels} at once (see Algorithm~\ref{ND++}).

Experiments show that the above formulation works well for simple classification tasks that do not require scale invariance (see results in Sec. \ref{sec:ablation}). However, it may not work well for some other tasks such as object detection and instance segmentation where scale invariance plays an important role. To generalize to these tasks, we incorporate local statistics to remove the scale in each individual sample. This simple strategy is inspired by the biological observation that \textit{animal visual systems also use two sets of statistics for visual analysis} \cite{hubel1962receptive,bear2020neuroscience}. The retinal light reflex~\cite{bear2020neuroscience} suggests local statistics is used to adjust the scale of the signal. For a fully-connected layer, local statistics are calculated from each row of the data matrix $X$. For a convolution/correlation layer, local statistics can be calculated for one or more rows of the data matrix or even the full feature tensor at each layer. We have found empirically that the last option works well and does not introduce an excessive computational cost. At each layer we standardize each sample tensor interchangeably with $\mu,\sigma$~\cite{ba2016layer}, or with mean $l_1$ norm $E(|x|)$ if we just consider the scale. Combined with the linear transform weights, our proposed feature transform is:

\begin{equation}
y=S\cdot X\cdot D \cdot w.
\label{transform}
\end{equation}
Here $S$ is the standardization operating on the \textit{rows} of $X$, a diagonal matrix ($\frac{1}{\sigma_i}$) if we only consider the scaling or augmented with an extra column ($-\frac{\mu_i}{\sigma_i}$) if we also consider the bias, and $D$ is the decorrelation operating on the \textit{columns} of $S\cdot X$. The weights are trained in the transformed space based on the uncorrelated features $S\cdot X\cdot D$. 

Our formulation transforms both the rows and columns of $X$ without introducing redundancy. Normalization using local statistics enforces scale invariance, aligning features of different scales. Normalization with global statistics enforces $GL(n)$-invariance, leading to faster convergence.

\section{Implementation Details}

\subsection{Cross-GPU Synchronization}

To leverage the latest advancements in hardware development, we implement cross-GPU synchronization to improve the quality of the estimates (see ~\cite{pan2019switchable} for another independent implementation). At each layer, the covariance matrix is computed on each GPU and synchronized across all the GPUs. On an 8-GPU machine, the synchronization cost is negligible. This implementation allows us to collect reliable statistics throughout the training for all practical batch sizes. In our development, we have consistently achieved satisfactory results using per-GPU batch sizes ranging from from $2$ to $1024$ in an $8$-GPU training environment. 

\subsection{Acceleration Techniques}
\label{sec:acceler}
We use various techniques to simplify computation and reduce the complexity of the proposed algorithm to a small fraction ($\sim 5\%$) of the cost of a convolution layer (Supp. Table~\ref{tab:runtime}). The memory accessing time of extracting the data matrix from a convolution/correlation layer in existing packages takes significantly more time than the underlying convolution operation. We adopt a $3\times$-subsampling for ImageNet scaled images and $5\times$-subsampling for MS COCO scaled images. The data and covariance matrices are computed using tensor patches sampled at the strided locations, reducing the construction cost by $9\times$ and $25\times$ respectively. Since the involved number of pixels is usually more than enough, this strategy maintains accuracy while keeping the computational footprint low. When the covariance matrix becomes too large, we divide the columns into blocks and decorrelate between the blocks~\cite{DBLP:journals/corr/abs-1708-00631,huang2018decorrelated}. In our experiments, we set the number of blocks $B=256$ for fully-connected layers and $B=64\times k^2$ for convolution/correlation layers.

There are multiple ways~\cite{huang2018decorrelated,DBLP:journals/corr/abs-1809-08625,huang2019iterative,Ye2020Network} to calculate the principal inverse square root of a matrix but algorithms that are both efficient and stable are scarce. Since explicit eigenvalue decomposition is slow, Newton-Schultz iteration becomes a faster alternative. However, the vanilla Newton-Schultz method is unstable under finite arithmetic~\cite{Ye2020Network}(Fig. 7), so we adopt a coupled inverse Newton iteration method that is numerically stable and works well under finite precision arithmetic~\cite{guo2006schur}. Starting with $X_0=I$, $M_0=Cov$, the coupled inverse Newton iteration calculates $X_{k+1}=X_{k}\frac{(3I-M_k)}{2}$, $M_{k+1}=(\frac{3I-M_k}{2})^2M_{k}$ and produces $X_k \xrightarrow[]{}Cov^{-\frac{1}{2}}$. Empirically, using $5$ iterations yields good results. We also add a small diagonal matrix $\epsilon I$ to the covariance matrix to avoid rank-deficiency. 

Note that for a large data matrix $X$, decorrelating columns of $X$ can require excessive computation. Previous applications are usually restricted to a small number of layers~\cite{huang2018decorrelated,huang2019iterative,pan2019switchable}, which adds to the network design complexity. We adopt a universal design and avoid excessive computation by reordering the computation by transforming the model weights instead, $y=(S\cdot X) \cdot (D \cdot w)$ (see Algorithm~\ref{ND++}). 


\section{Experiments}
We refer to our improved implementation as Network Deconvolution++ (ND++). ND++ introduces scale invariance, generalizes the $GL(n)$-invariance to all three common linear transform layers (convolution/correlation/linear) in modern architectures, and uses cross-GPU synchronization to allow reliable training at different scales. 
In the following experiments, all linear transform layers are enhanced with ND++. 
In fine-tuning experiments, the pretrained backbone network is replaced with a backbone pretrained with ND++. For a fair comparison, we have also experimented with training from scratch to verify the gain is not only from the improved backbone. Standard stochastic gradient descent is used for all experiments. We continue to use learning rate decay in the face of inherent non-linearity and mini-batch training.

\subsection{Image Classification}

\begin{table}[]
\centering
\begin{tabular}{l|r|r}
\multicolumn{1}{c|}{Network} & \multicolumn{1}{c|}{SyncBN} & \multicolumn{1}{c}{ND++} \\ \hline
VGG-11                       & 71.11                 & \textbf{72.24}                  \\
ResNet-50                    & 76.25                 & \textbf{77.95}                  \\
ResNet-101                    & 77.37                 & \textbf{79.40}                  \\
DenseNet-121                 & 74.65                 & \textbf{76.11}
\end{tabular}
\caption[ImageNet]{Top-1 Validation Accuracy on the ImageNet Dataset. ND++ also surpasses the top-1 accuracy rates of the deeper networks in the model zoo: VGG-13: $71.55\%, $ResNet-101: $77.37\%, $ResNet-152: $78.32\%, $DenseNet-169: $76.00\%$.}
\label{tab:ImageNet}
\end{table}

We demonstrate improved training on three popular CNN architectures (VGG, ResNet, DenseNet) at three scales (10/50/100 layers). With ND++, we have seamlessly increased the training to batch size $2048$, eight times larger than the model zoo default setting of $256$. We train in one-eighth the number of iterations but produce superior models (Table~\ref{tab:ImageNet}). All three network architectures, VGG-11/ResNet-50/ResNet-101/DenseNet-121, surpass \textit{their deeper counterparts} in the model zoo after standard 90-epoch training with cosine learning rate decay. Using the official PyTorch recipe, it takes less than one day to train on a machine with 8 Nvidia A100 GPUs. 
On the popular ResNet-50/101 network, the \bm{$77.95\%$}/\bm{$79.40\%$} top-1 accuracy we reach is among the highest numbers reported when training for 90 epochs (Fig.~\ref{fig:frontpage}(c)). 

\subsection{Object Detection}

\begin{table}[]
\centering
\begin{tabular}{l|r|r|r}
     & \multicolumn{1}{c|}{MB}         & \multicolumn{1}{c|}{MLPerf}     & \multicolumn{1}{c}{Detectron2}  \\ \hline
ND++ & \textbf{37.36} & \textbf{37.37} & \textbf{39.01} \\
BN   & 36.78                           & 36.35                           & 37.9                            \\
GN   & 36.04                           & 35.9                            & 38.54                          
\end{tabular}%
\caption[FasterRCNN]{Bounding box AP of three Faster R-CNN implementations on the COCO 2017 dataset.}
\label{tab:FasterRCNN}
\end{table}

\begin{table}[]
\resizebox{0.49\textwidth}{!}{
\centering
\begin{tabular}{l|rr|rr|rr}
     & \multicolumn{2}{c|}{MB}                                            & \multicolumn{2}{c|}{MLPerf}                                        & \multicolumn{2}{c}{Detectron2}                                   \\
     & \multicolumn{1}{c}{$AP^{bbox}$} & \multicolumn{1}{c|}{$AP^{mask}$} & \multicolumn{1}{c}{$AP^{bbox}$} & \multicolumn{1}{c|}{$AP^{mask}$} & \multicolumn{1}{c}{$AP^{bbox}$} & \multicolumn{1}{c}{$AP^{mask}$} \\ \hline
ND++ & \textbf{38.62}                  & \textbf{35.65}                   & \textbf{38.36}                  & \textbf{35.07}                   & \textbf{39.91}                  & \textbf{36.87}                  \\
BN   & 37.67                           & 34.28                            & 37.14                           & 33.97                            & 38.6                            & 35.2                            \\
GN   & 37.82                           & 34.75                            & 36.32                           & 33.52                            & 38.96                           & 35.93                          
\end{tabular}%
}
\caption[MaskRCNN]{Bounding box and mask AP of three Mask R-CNN implementations on the COCO 2017 dataset.}
\label{tab:MaskRCNN}
\end{table}

We test ND++ on Faster R-CNN and Mask R-CNN, milestone object detectors from three major benchmarks, \textit{maskrcnn-benchmark}(MB)
, \textit{MLPerf}
, and \textit{Detectron2}. We use ResNet-50 with FPN as backbones and two fully-connected layers in the box heads. In Mask R-CNN, the mask head contains convolution and correlation layers. ND++ is used to enhance all layers. We report our numbers and evaluation curves on the COCO 2017 dataset (Tables~\ref{tab:FasterRCNN},~\ref{tab:MaskRCNN}, Fig.~\ref{fig:frontpage}(a,b)). Results with ResNet-101 can be found in Supp. Sec.~\ref{sec:OD}, Supp. Tab.~\ref{tab:detectron2}.

\begin{table}[]
\centering
\begin{tabular}{l|lll}
     & \multicolumn{1}{c}{LR} & \multicolumn{1}{c}{$AP^{bbox}$} & \multicolumn{1}{c}{$AP^{mask}$} \\ \hline
ND++ & 0.02                   & 37.81                       & 34.87                       \\
GN   & 0.02                   & 37.76                       & 34.8                        \\ \hline
ND++ & 0.1                    & \textbf{38.86}              & \textbf{35.56}              \\
GN   & 0.1                    & 35.21                       & 33.66
\end{tabular}
\caption[scratch]{Mask R-CNN trained from scratch on the COCO 2017 dataset.}
\label{tab:scratch}
\end{table}

On all three benchmarks and with both Faster R-CNN (Table~\ref{tab:FasterRCNN}) and Mask R-CNN (Table~\ref{tab:MaskRCNN}), ND++ consistently outperforms baselines with frozen batch normalization and group normalization~\cite{GroupNorm} in terms of Average Precision(AP) (Fig.~\ref{fig:frontpage}(a,b)). 
When training from scratch, ND++ benefits from increase the learning rate to $0.1$ while group normalization ends up with worse results (Table~\ref{tab:scratch}, Supp. Fig.~\ref{fig:maskrcnn_scratch}).

\begin{table}[]
\begin{tabular}{l|rr|rr}
      & \multicolumn{1}{c}{LAMB}    & \multicolumn{1}{c|}{MegDet}  & \multicolumn{2}{c}{ND++}                                 \\
Batch & \multicolumn{1}{c}{$AP^{bbox}$} & \multicolumn{1}{c|}{$AP^{bbox}$} & \multicolumn{1}{c}{$AP^{bbox}$} & \multicolumn{1}{c}{$AP^{mask}$} \\ \hline
128   & 36.7                        & 37.7                         & \textbf{38.61}              & 35.54                       \\
256   & 36.7                        & 37.7                         & \textbf{38.40}               & 35.26                       \\
512   & 36.5                        & -                            & \textbf{37.45}              & 34.5
\end{tabular}
\caption[large batch]{Large-scale training performance of Mask R-CNN using ND++. Our reported numbers are based on the NVIDIA implementation. ND++ significantly surpasses MegDet~\cite{DBLP:journals/corr/abs-1711-07240} and LAMB~\cite{wanglarge} that utilize SyncBN, warmup and optimizer change.}
\label{tab:large_batch}
\end{table}

Though researchers have encountered severe challenges in scaling up the training of Mask R-CNN~\cite{DBLP:journals/corr/abs-1711-07240,wanglarge}, we have been able to seamlessly increase the batch size and achieve good results without ad hoc techniques. Here we take the MLPerf implementation, remove the warmup stage, and fix the learning rate to 0.1 with a momentum of 0.9. We notice that frozen/synchronized batch normalization explodes at all scales without warmup while ND++ produces superior results. ND++ significantly surpasses the MLPerf accuracy goal\footnote{37.7 for bounding box AP and 33.9 for mask AP.} within the standard 12-epoch fine-tuning with batch sizes up to $256$, an order of magnitude larger than most existing baseline settings (Table~\ref{tab:large_batch}, Supp. Fig.~\ref{fig:maskrcnn_large_batch}). 

\subsection{Semantic Segmentation}


\begin{table}[]
\begin{tabular}{l|c|c}
Network                           & SyncBN  & ND++           \\ \hline
DLv3-RN-50 (scratch, 200 ep.)              & 69.30   & \textbf{72.69} \\
DLv3-RN-50 (finetune, 200 ep.)             & 75.70   & \textbf{77.50} \\
DLv3-RN-101 (finetune, 200 ep.)            & 77.28   & \textbf{79.23} \\
DLv3-RN-50 (finetune, 50 ep.)     & -       & 76.47          \\
DLv3-RN-50 (finetune, 500 ep.)    & 75.71   & -
\end{tabular}
\caption[Cityscapes]{Validation mIoU when fine-tuning and training from scratch on the Cityscapes dataset. DLv3 stands for DeepLabv3 and RN stands for ResNet.}\label{tab:Cityscapes}
\end{table}

To demonstrate the usefulness of ND++ for semantic segmentation, we add ND++ layers to the DeepLabv3 architecture~\cite{chen2017rethinking} with both ResNet-50 and ResNet-101 backbones and test the performance on the Cityscapes dataset~\cite{cordts2016cityscapes}. We use a base resolution of 1024 for the images. 
We train for 200 epochs for all experiments unless stated otherwise. We perform fine-tuning experiments with a ResNet-50 backbone pretrained on ImageNet an initial learning rates of 0.01 and 0.1 and momentum 0.9 for both ND++ and the SyncBN baseline, and we report the results with learning rate 0.01 for SyncBN and 0.1 for ND++ in Table~\ref{tab:Cityscapes} since these produced the best results for each network configuration. Our ResNet-50 model with ND++ achieves an mIoU of 77.50 on Cityscapes, comparable to the ResNet-101 model from the original paper with 77.82 mIoU for single-scale evaluation~\cite{chen2017rethinking}.

ND++ substantially improves over the synchronized batch norm baseline, and the acceleration is especially apparent early in training. For example, it takes less than 50 epochs for a network equipped with ND++ to beat the performance of a network trained using SyncBN for 500 epochs (Supp. Fig.~\ref{fig:one-tenth} and Table~\ref{tab:Cityscapes}) if we fine-tune using a pretrained ResNet-50 backbone.

\subsection{Experiments with Learning Rate $1.0$}

\begin{table}[]
\resizebox{0.49\textwidth}{!}{
\begin{tabular}{c|cc|c}
ResNet-50 & \multicolumn{2}{c|}{Mask R-CNN (MLPerf)} & DeepLabv3-RN-50 \\
top-1     & AP bbox            & AP mask           & mIoU (Scratch)      \\ \hline
77.66     & 38.26              & 35.41             & 71.64
\end{tabular}
}
\caption[LR1]{Experiments with an initial learning rate $1.0$ and no momentum.}
\label{tab:LR1}
\end{table}

Existing deep neural network models are trained with stochastic gradient descent algorithm variants with momentum~\cite{sutskever2013importance}. Although the momentum term usually reduces noise and accelerates convergence, we notice that with ND++, many networks train well with a initial learning rate $1.0$ and without the use of the momentum term, presumably thanks to the $GL(n)$-invariance property (Table~\ref{tab:LR1}).

\subsection{Training a Vision Transformer Model}
The techniques discussed in this paper naturally generalize to emerging architectures. To demonstrate, we compare a $10,000$-step training using ND++ and compare with ND and the standard training of a vision transformer (ViT-B\_16) on the CIFAR-10/100 datasets. Mostly following the original setting, we resize images to 224, set the learning rate of SGD to 0.1 and the weight decay to $0.0001$, and use a batch size of 512 when training from scratch\footnote{The Vision transformer is accessed from: https://github.com/jeonsworld/ViT-pytorch
}. Note that even though fine-tuning Transformers has shown promising results on various language and vision tasks, training transformers from scratch leaves large room for improvements. When training from scratch on the CIFAR-10 dataset, the baseline $10,000$-step (102 epochs) training yields an accuracy of only $70\%$. We drop in our modifications in every linear layer of the network and remove the original LayerNorms in the network. Interestingly, we observe $GL(n)$-invariance suffices and significantly improves the testing accuracy curve. Adding scale invariance further improve the results (Fig.~\ref{fig:frontpage}(d)) to $80\%$. More training from scratch results for language tasks can be found in the supplementary materials. The wall-time of the baseline/ND/ND++ is 180/200/220 minutes, respectively, when run on 8 GPUs.

\subsection{Ablation Studies}
\label{sec:ablation}

\begin{figure}[hbt!]
\centering
\subfigure[]{\includegraphics[width=.32\columnwidth]{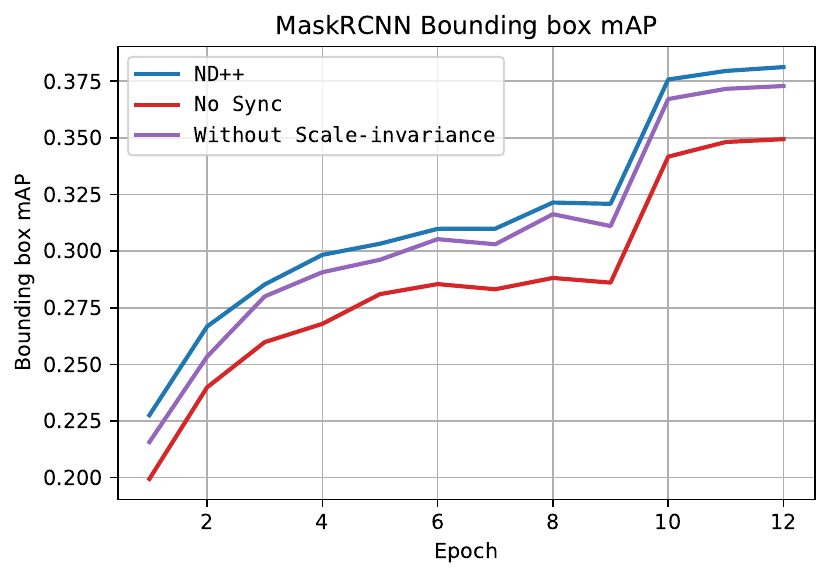}}
\subfigure[]{\includegraphics[width=.32\columnwidth]{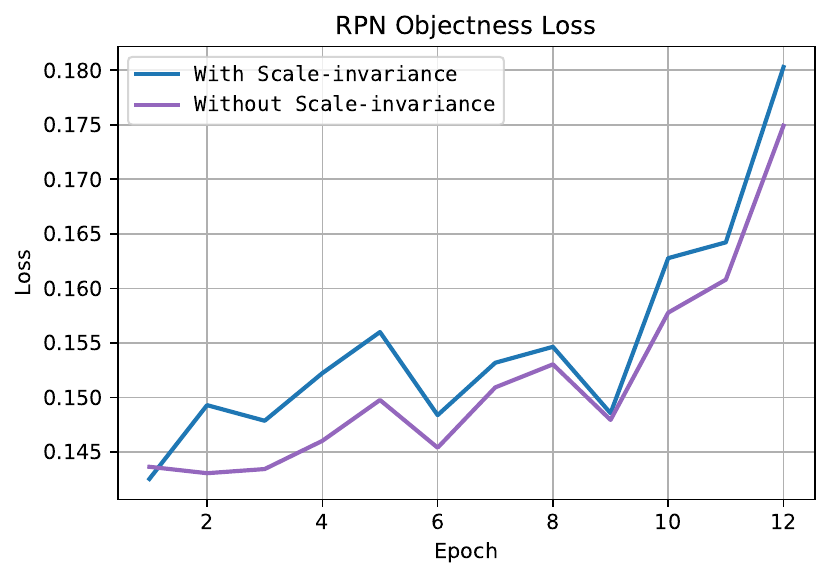}}
\subfigure[]{\includegraphics[width=.32\columnwidth]{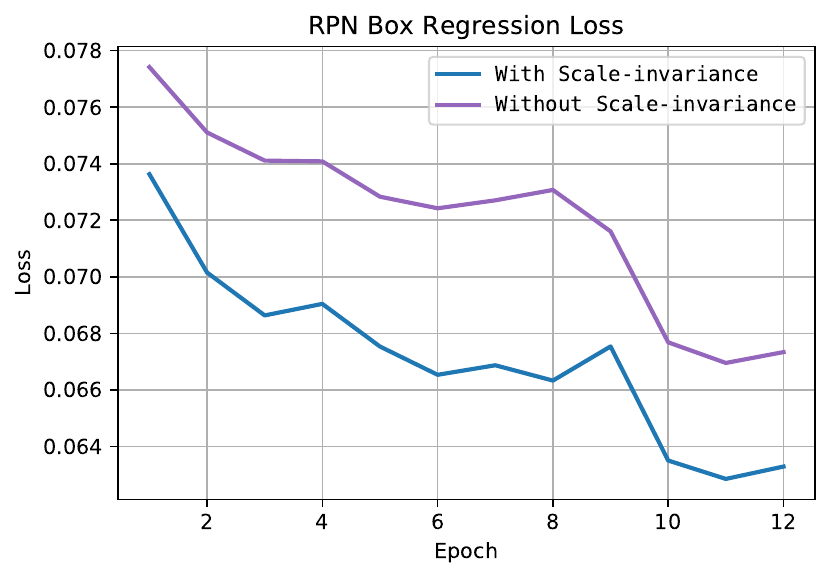}}

\caption[ablation]{Ablation study with the Mask R-CNN architecture.}
\label{fig:maskrcnn_ablation}
\end{figure}

Cross-GPU synchronization and scale invariance produce larger improvements in the more advanced two-stage detector Mask R-CNN as indicated by ablation experiments on the MLPerf benchmark (Fig.~\ref{fig:maskrcnn_ablation} (a)). To further study the effect of enforcing scale invariance, we separately plot the objectness classification and regression losses for the region proposal network (RPN) (Fig.~\ref{fig:maskrcnn_ablation} (b,c)). Without scale invariance, the object classification loss remains lower throughout the training, but the bounding box regression loss is higher. The inaccurate proposals likely pull down the final results (Fig.~\ref{fig:maskrcnn_ablation} (a) the purple curve). Further studies can be found in the experiment details in the supplementary materials, Supp. Sec~\ref{suppexp}.

\section{Discussion}

The transform we propose has simple and intuitive geometric meanings. To achieve scale invariance, each sample is stretched individually using local statistics. The global geometry of the loss landscape, however, depends on the full collection of data. We therefore utilize the global distribution of data to find a unique optimal feature transform and achieve $GL(n)$-invariance. The unique gradient direction in this space corresponds to the optimal direction in the linear case and leads to significantly accelerated training of deep neural networks.

{\small
\bibstyle{aaai22}
\bibliography{egbib.bib}
}

\newpage
 
\appendix

\section{Source Code}

We present a PyTorch-like code snippet showing our modifications to a linear layer.

The full source code can be accessed from: https://github.com/amazon-research/network-deconvolution-pp



\begin{figure}
\includegraphics[width=.99\columnwidth]{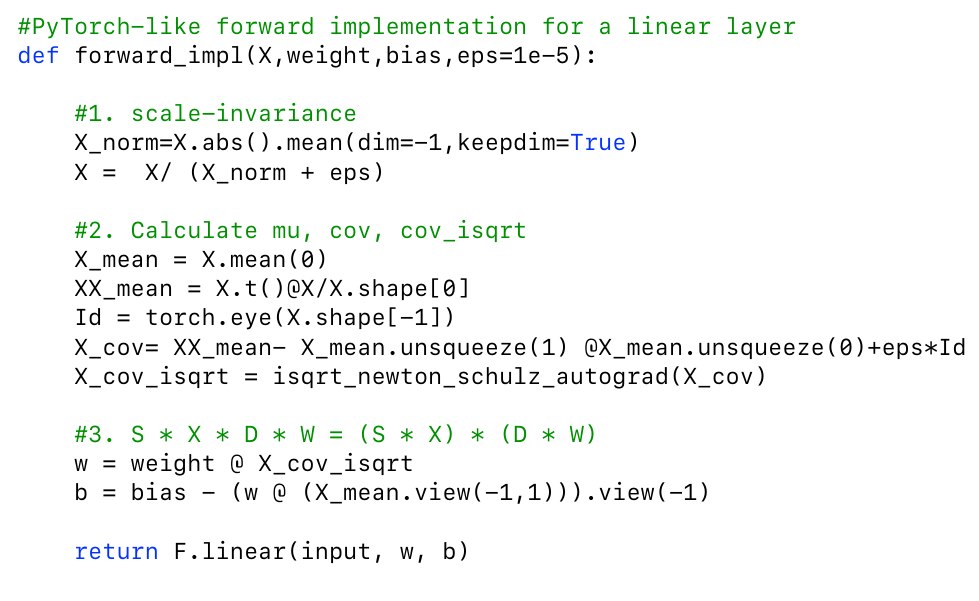}
\caption[sample code]{Reference modifications to a linear layer.}\label{fig:samplecode}
\end{figure}

\section{Cat Receptive Fields}

\begin{figure}[hbt!]
\centering
\includegraphics[width=.99\linewidth]{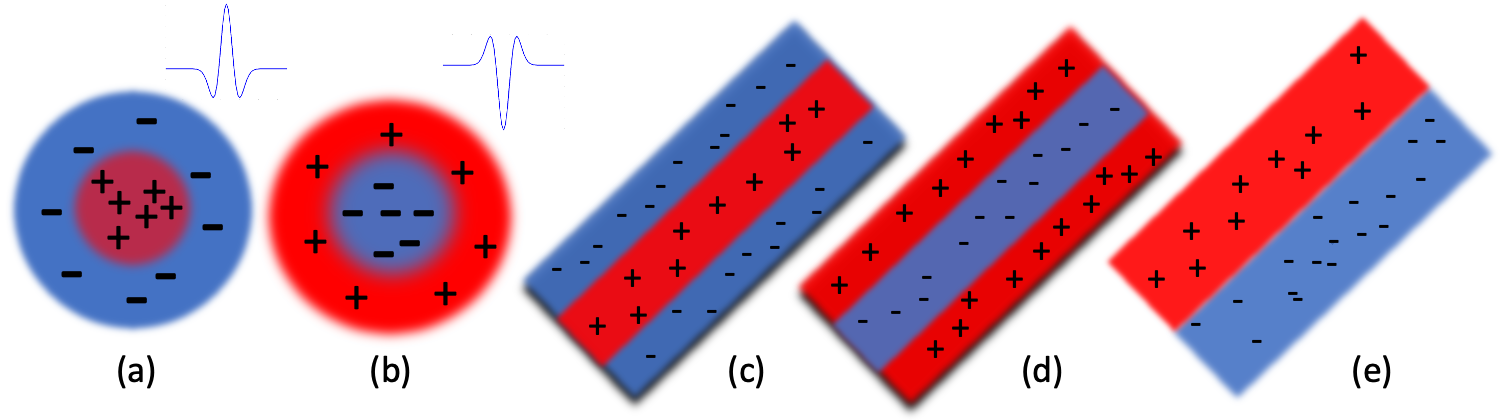}
\caption[correction methods]{Receptive fields in the cat retina. Reproduced from~\cite{hubel1962receptive}, Fig. 2.}\label{fig:hubel-wiesel}
\end{figure}

\section{Further Discussion}
~\label{sec:clarification}
We can compute the covariance matrix $Cov_1=X_1'X_1$ from the data matrix $X_1$ of feature channel 1 to remove the correlation in nearby pixels. For multiple channels we construct $X=[X_1,...,X_C]$ and compute a larger covariance matrix $Cov=X'X$. The diagonal blocks $X_i'X_i$ of this matrix measure the correlation of nearby pixels in each channel. The off-diagonal blocks $X_i'X_j$ measures the (pixel-wise) correlation across different channels ($i$ and $j$).  The transform we propose has a simple form for each sample: $z(x)=\Sigma^{-0.5}(s(x)-\mu)$, where $s(x)$ is a simple scaling for each sample. $\mu,\Sigma$ are computed from a batch.

Our formulation in Sec.~\ref{sec:invariance} is based on the mathematical convolution. In most deep learning packages, the `convolution' is implemented without flipping the kernel and is therefore in fact the `correlation' operation. This misnaming does not affect the properties that we discuss throughout the paper.

\subsection{Scale Invariance}

We present our argument on scale invariance based on linear transforms without the bias term. Though the bias term can potentially mitigate the problem, it cannot introduce scale invariance. In classification problems, the softmax function introduces scale invariance to the logits. This is why we see smaller improvement from scale invariance compared to object detection.

\subsection{GL(n)-Invariance}
\begin{figure}[hbt!]
\centering
\includegraphics[width=.99\linewidth]{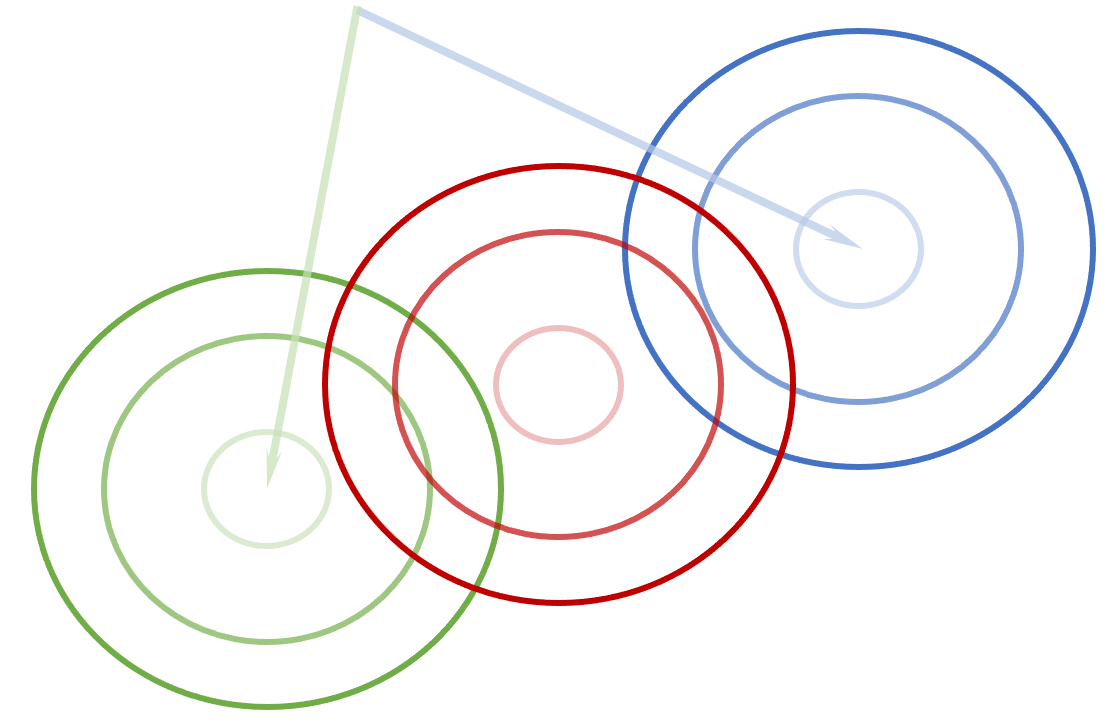}
\caption[GL(n)-invariance]{During the training, every minibatch has an individual loss landscape (visualized in green and blue). These loss landscapes are used during the training. During the testing, the global loss landscape (visualized in red) is used. This picture gives us intuition about the training/testing inconsistency and how issue should be properly addressed, as discussed in the text.}\label{fig:GLn_inv}
\end{figure}

Here we present a global picture of $GL(n)$-invariance. The ideal solution is to generate the global loss landscape with the whole dataset (Fig.~\ref{fig:GLn_inv}(red)). Due to hardware limitations, training is conducted in minibatches. During training, different batches of data give rise to different loss landscapes (Fig.~\ref{fig:GLn_inv}(green and blue)). These landscapes from sampled data generate noisy gradient directions that do not lead to the optimal solution for the whole dataset (Fig.~\ref{fig:GLn_inv} (red)). Gradient descent follows these noisy gradients in an online fashion to update weights. Due to this approximation, it is beneficial to use learning rate decay or increase the batch size~\cite{l.2018dont} towards the end of the training. However, using the running average covariance matrix to transform a batch of data yields poor performance. This incorrect transform can result in a heavily curved loss landscape. Therefore, the covariance needs to be calculated from the same batch of data to avoid divergence. The evaluation, however, is based on the global loss landscape over the whole dataset (Fig.~\ref{fig:GLn_inv}(red)). The global transform can be achieved through the use of running average statistics. Methods that use global statistics such as the batch normalization are known to have this inconsistency in training and testing. The inconsistency can be alleviated by reducing the learning rate towards the end of the training or calculating the global statistics more accurately. One can also transform back to the original space (with the help of $Cov^\frac{1}{2}$) to eliminate the inconsistency. The methods that only utilize local statistics~\cite{ba2016layer,GroupNorm,singh2020filter} to achieve consistency in training and testing in fact use a suboptimal geometry in the training and therefore have slower convergence.

\section{Complexity Analysis}
~\label{sec:complexity}
We derive the complexity of our proposed transform in a matrix multiplication layer $Y=(S\cdot X)\cdot ({Cov}^{-\frac{1}{2}} \cdot W)$. The complexity analysis can be easily generalized to convolution and correlation layers~\cite{Ye2020Network}. $W$ is the weight matrix and has dimensions ${Ch}_{in}\times{Ch}_{out}$. $S \cdot X$ introduces scale invariance. Note that $S$ is either diagonal or augmented with an extra column if we include bias. The complexity is $O(N\times {Ch}_{in})$ , which is linear to the scale of $X$. We put the emphasis on the computation of ${Cov}^{-\frac{1}{2}} \cdot W$. We first assume $X$ has been reshaped to $N\cdot \frac{{Ch}_{in}}{B} \times B$. The covariance matrix is calculated based on the reshaped $X$ and has a size of $B \times B$. The computation of the covariance matrix has complexity $O(B \cdot \frac{N \cdot {Ch}_{in}}{B} \cdot B  )=O(N\cdot {Ch}_{in} \cdot B )$. Coupled inverse Newton-Schulz iterations ($5$-iterations) are therefore conducted on matrices of size $B \times B$. Solving for the inverse square root takes $O(B^3)$. Next we assume $W$ has been reshaped from ${Ch}_{in}\times {Ch}_{out}$ to $B\times (\frac{{Ch}_{in}\cdot {Ch}_{out}}{B})$, weight transform $Cov^{-\frac{1}{2}}\cdot W$ is a simple matrix multiplication with complextity $O(B^2 \cdot \frac{{Ch}_{in}\cdot {Ch}_{out}}{B} )=O(B \cdot {Ch}_{in}\cdot {Ch}_{out} )$. The overall complexity is $O( N\cdot {Ch}_{in} \cdot B +B^3+ B \cdot {Ch}_{in} \cdot {Ch}_{out})=O(( N\cdot {Ch}_{in}+B^2+{Ch}_{in} \cdot {Ch}_{out})\cdot B)$. This can be a small fraction of the cost of the matrix multiplication operation that has a complexity of $O(N \times {Ch}_{in}\times {Ch}_{out})$ if we use a small $B$. The simple derivation above can be easily generalize to the case of convolution and correlation layers. Note that $N$ is approximately equal to the number of pixels at each layer.

\section{Runtime Analysis}
\begin{table}[]
\begin{tabular}{l|rrrrr}
$Kernel$      & 7     & 3    & 3     & 1    & 1    \\
$Stride$      & 2     & 1    & 1     & 1    & 2    \\
$Ch_{in}$      & 3     & 64   & 256   & 512  & 2048 \\
$Ch_{out}$     & 64    & 64   & 256   & 128  & 1024 \\
$H$           & 800   & 200  & 200   & 100  & 50   \\
$W$           & 1333  & 333  & 333   & 166  & 83   \\ \hline
$\frac{x}{\sigma_1}$      & 0.75  & 0.99 & 3.84  & 1.94 & 1.95 \\
$\frac{x-\mu}{\sigma_2}$   & 1.81  & 2.39 & 9.38  & 4.74 & 4.73 \\ \hline
Get $X$ & 0.72  & 0.93 & 6.22  & 2.81 & 2.93 \\
$Cov=\frac{1}{N}X^tX$         & 1.43  & 0.7  & 2.75  & 1.36 & 1.4  \\
$D=Cov^{-\frac{1}{2}}$       & 0     & 0.14 & 0.13  & 0.13 & 0.13 \\
$D \cdot W$          & 0     & 0    & 0.02  & 0.02 & 0.84 \\ \hline
$Conv$        & 30.16 & 3.4  & 30.56 & 7.31 & 46.6
\end{tabular}
\caption[runtime]{CPU runtime (in seconds) for a few representative settings in a Mask R-CNN network. The top rows in the first block show the layout in the network, the rows in the second block shows the runtime for two local normalization methods, the rows in the third block show the runtime for the $GL(n)$-invariance. The last row shows the runtime of the convolution layer.}\label{tab:runtime}
\end{table}

Here we report the CPU time for a few representative layers in a Mask R-CNN network to verify the claim about the complexity of our proposed algorithm (Table~\ref{tab:runtime}). The batch size is set to $16$, the subsampling stride is set to $5$ to compute the covariance matrix, and timing is averaged over $10$ batches. Note that the overall computation required for introducing the $GL(n)$-invariance is comparable to introducing scale invariance and is usually a small fraction of the time taken by the convolution layer. Moreover, extracting the data matrix, which takes the most time, \textit{shares the same computation as the underlying convolution}, and therefore would not be required with proper optimization. However, this sharing mechanism is not currently supported in common software packages. This lack of low-level optimization leads to extended wall time in our experiments.

\section{Experiment Details}
\label{suppexp}

\subsection{Image Classification}
For the ImageNet classification experiments, we tested three weight decay levels (1e-4/5e-4/1e-3) and report the highest accuracy for both the synchronized batch normalization baseline and ND++. We have also tested batch normalization without synchronization and find the synchronized version usually gives a better result; we therefore use it as the baseline. The baseline model achieves the reported accuracy with weight decay factors 1e-4/5e-4/5e-4 respectively for VGG-11/Resnet-50/DenseNet-121. Correlated features create pathological regions that prevents the usage of large weight decay (Fig~\ref{fig:correction}). With ND++, larger weight decay factors can be used and the optimal factors are 5e-4/1e-3/1e-3. For DenseNet, the baseline model takes more GPU RAM than the machine can hold and the result with batch size $1024$ is reported here to compare with our batch size $2048$ result. Due to the latency of the customized implementation, the wall time of ND++ is longer but still acceptable. By using the large batch size, the number of calls to the slow functions is reduced, and we finish the training in a shorter time compared to the original batch size of 256. With batch size 256, the baseline with SyncBN took 29 hours to train the ResNet-50. With batch size 2048, the complete version of ND++ finishes in the same time with superior accuracy. If remove scale invariance, ND++ finishes in 24 hours with a negligible drop in accuracy. At batch size 2048, ND++ without scale invariance takes 23/24/36 hours for VGG-11/ResNet-50/DenseNet-121 while the SyncBN baselines take 18/19/21 hours. Enforcing scale invariance slows training by 20\% with little change in performance. The experiments are conducted on an AWS EC2 P3dn.24x instance. Note that the reported timing allows plenty of room for future improvements: besides sharing the extraction of the data matrix, layerwise standardization can be accelerated with sampling techniques.

For ImageNet training, we see a tiny improvement by enforcing scale invariance, while synchronization does not seem to give us a significant gain due to the per GPU batch size of $256$. If we remove both synchronization and scale invariance, the top-1 accuracy of ResNet-50 drops slightly to $77.80\%$.

\subsection{Object Detection}
\label{sec:OD}

\begin{table}[]
\resizebox{0.45\textwidth}{!}{%
\begin{tabular}{l|rr|rr}
                             & \multicolumn{2}{c|}{Model Zoo Baseline}                    & \multicolumn{2}{c}{ND++}                                 \\
                             & \multicolumn{1}{c}{$AP^{bbox}$} & \multicolumn{1}{c|}{$AP^{mask}$} & \multicolumn{1}{c}{$AP^{bbox}$} & \multicolumn{1}{c}{$AP^{mask}$} \\ \hline
Faster R-CNN R101-FPN        & 42                          &                              & 43.24                       &                             \\
Mask R-CNN R101-FPN          & 42.9                        & 38.6                         & 43.86                       & 39.96                       \\
Mask R-CNN R50-FPN (scratch) & 39.9                        & 36.6                         & 41.12                       & 37.36                      
\end{tabular}%
}
\caption[Results on Detectron2]{Comparison with Detectron2 3x baselines found in the model zoo. For our R101 results, we increase the batch size by 3x instead of the training steps.}
\label{tab:detectron2}

\end{table}

\begin{figure}[hbt!]
\centering
\subfigure[]{\includegraphics[width=.49\columnwidth]{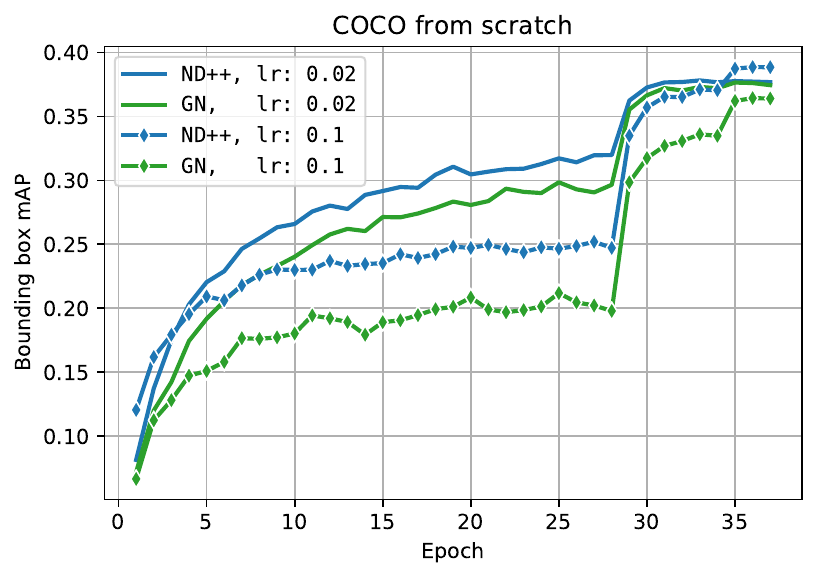}}
\subfigure[]{\includegraphics[width=.49\columnwidth]{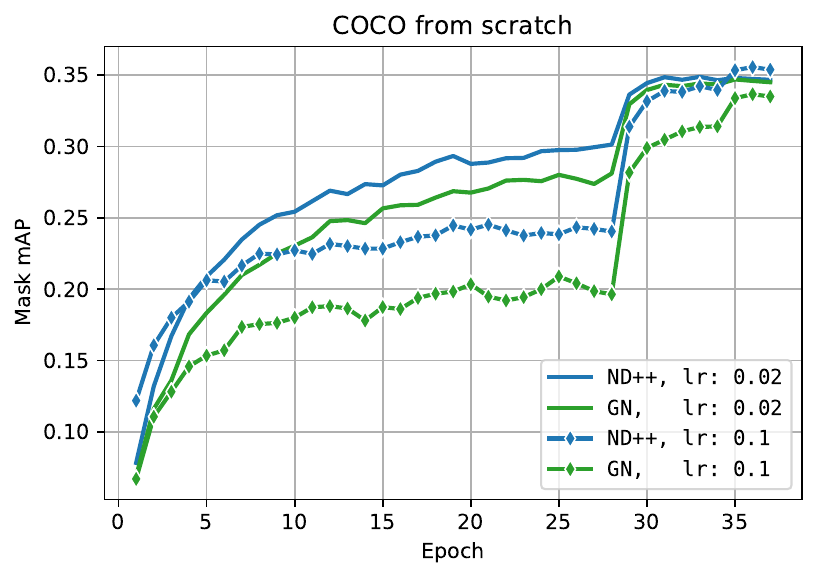}}
\caption[maskrcnn scratch]{Bounding box and mask AP curves during validation when training the Mask R-CNN network from scratch. Note that with the more standard learning rate of 0.1, ND++ produces superior results compared to the GN baseline.}
\label{fig:maskrcnn_scratch}
\end{figure}

\begin{figure}[hbt!]
\centering
\subfigure[]{\includegraphics[width=.49\columnwidth]{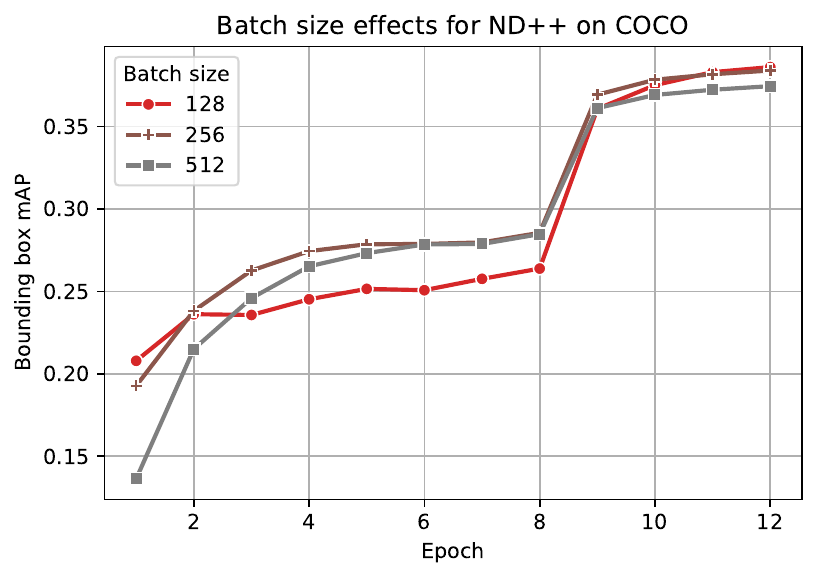}}
\subfigure[]{\includegraphics[width=.49\columnwidth]{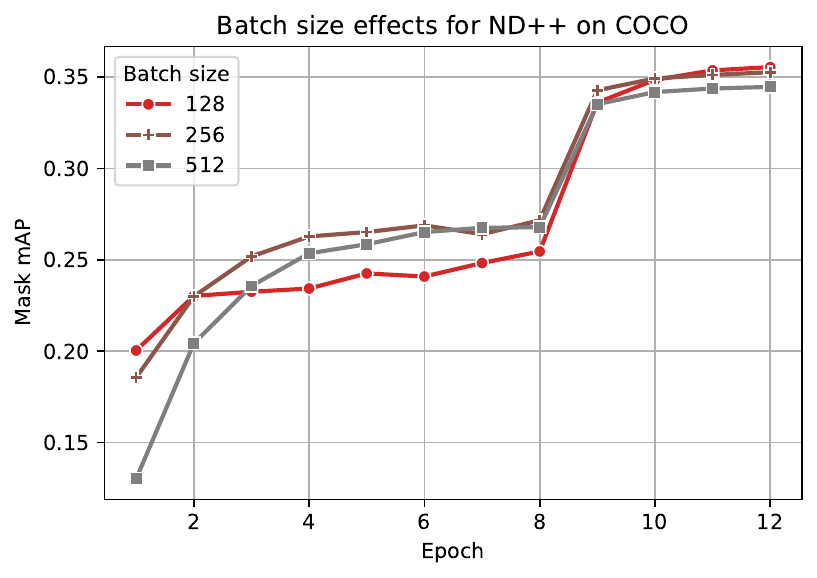}}
\caption[maskrcnn large batch]{Bounding box and mask AP curves during validation when using large batch sizes to fine-tune the Mask R-CNN network.}
\label{fig:maskrcnn_large_batch}
\end{figure}

\begin{table}[]
\centering
\begin{tabular}{l|rr}
          & \multicolumn{1}{c}{AP bbox} & \multicolumn{1}{c}{AP mask} \\ \hline
Frozen BN & NaN                         & NaN                         \\
GN        & 33.46                       & 31.09                       \\
ND++      & 38.02                       & 35.09                      
\end{tabular}
\caption[maskrcnn LR 1]{Results of using an initial starting learning rate 1.0 to train a Mask R-CNN network.}
\label{tab:maskrcnn_lr1}
\end{table}

For object detection, we report additionally results of training Mask R-CNN on Detectron2. We attach figures of training from scratch with different learning rates (Supp. Fig.~\ref{fig:maskrcnn_scratch}) and fine-tuning with large batch sizes (Supp. Fig.~\ref{fig:maskrcnn_large_batch}). In most experiments we achieve impressive performance with ND++ while using a learning rate of 0.1 and a momentum of 0.9. This combination has an effective learning rate of 1.0~\cite{li2020rethinking}. Our transformation can make better use of this optimal learning rate compared to existing methods. The partially corrected gradients with standardization have less optimal direction and less useful scale. Smaller learning rates need to be used to avoid divergence. 

In this task, the lack of support to our customized implementation becomes more apparent due to the small training batch and layerwise standardization. In the MLPerf implementation, training and validating Mask R-CNN for 12 epochs takes 5 hours for the frozen batch normalization baseline, 6 hours for the group normalization baseline, can take more than 10 hours for ND++ using on an AWS EC2 P4d.24x instance with 8 GPUs. We notice that the seemingly simple layerwise standardization takes a significant amount of time. The runtime is 14 hours if we remove both scale and bias (Box AP 38.4), 11 hours if we remove just the scale (Box AP 38.3), and 9 hours if we do not apply scale invariance at all (Box AP 37.5). The wall time can be significantly reduced if we increase the batch size and reduce the number of function calls. At batch size 64, our slowest run takes 11 hours (3 hours faster), reaching Box AP 38.7 and Mask AP 35.6, superior to a recent method that trains with twice as many steps~\cite{yan2020towards}. Besides optimizing these operations through low-level changes to the software package, we can also use standardization layers in only a subset of the network. Our experimentation during development indicates that applying standardization only in the RPN can maintain high-quality results. Here we report the numbers for the slow but universal design. For large batch experiments, we remove the warmup iterations and increase the weight decay factor for large batch sizes. For batch sizes of 128/256/512, the weight decay factors we use are 2.5e-4/5e-4/1e-3. We distribute the computation over 2/4/8 machines. For these experiments, slow inter-machine communication is the major bottleneck. We use the AWS Elastic Fiber Adaptor to speed up the inter-node communication \textit{without changing the code}. These multi-machine experiments (including training and evaluation) finish in 7/5/4 hours respectively.

We also tested using an initial learning rate of 1.0 to train the Mask R-CNN network under different normalization techniques. We notice the training with Frozen Batch Normalization explodes early in the training. Group Normalization finishes with significantly worse results. ND++ ends up with competitive results (Supp. Table~\ref{tab:maskrcnn_lr1}).

\subsection{Semantic Segmentation}

\begin{table}[]
  \centering
  \begin{tabular}{l|c}
    & Validation mIoU \\ \hline
    ND++                  & \textbf{72.71} \\
    ND++ w/o scale inv.   & 72.69 \\
    ND++ w/o sync         & 69.82 \\
    BN                    & 69.30 \\
  \end{tabular}
  \caption[SemSegAblation]{Ablation studies for DeepLabv3 with a ResNet-50 backbone on the Cityscapes dataset. We train from scratch for each condition for 200 epochs.}\label{tab:SemSegAblation}
\end{table}

For semantic segmentation, we use the $2975$ images with fine annotations and report results on the validation split with $500$ images on the Cityscapes dataset. In all experiments, we perform standard data augmentation. We perform distributed training on 8-GPU machines for 200 epochs with a minibatch size of 3. We reduce the learning rate on a default polynomial schedule with exponent 0.9 and use a weight decay of 0.0001.

\begin{figure}[hbt!]
\centering
\subfigure[]{\includegraphics[width=.49\columnwidth]{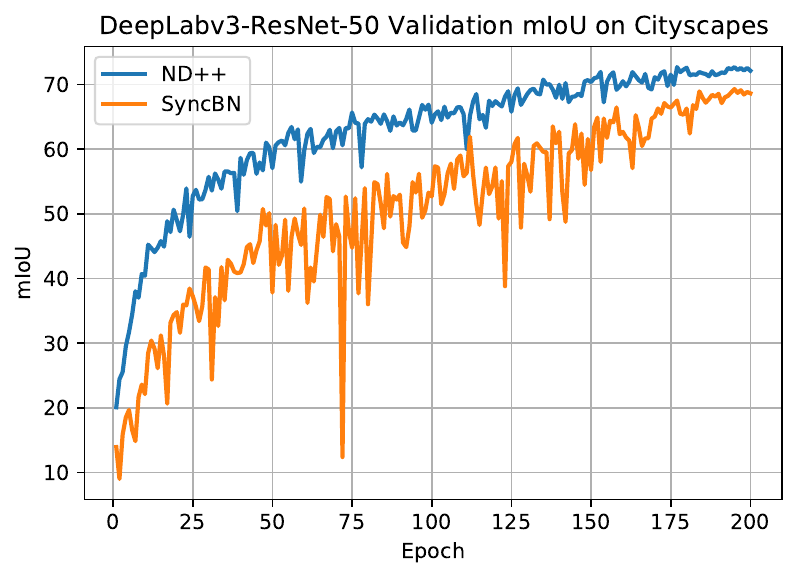}}
\subfigure[]{\includegraphics[width=.49\columnwidth]{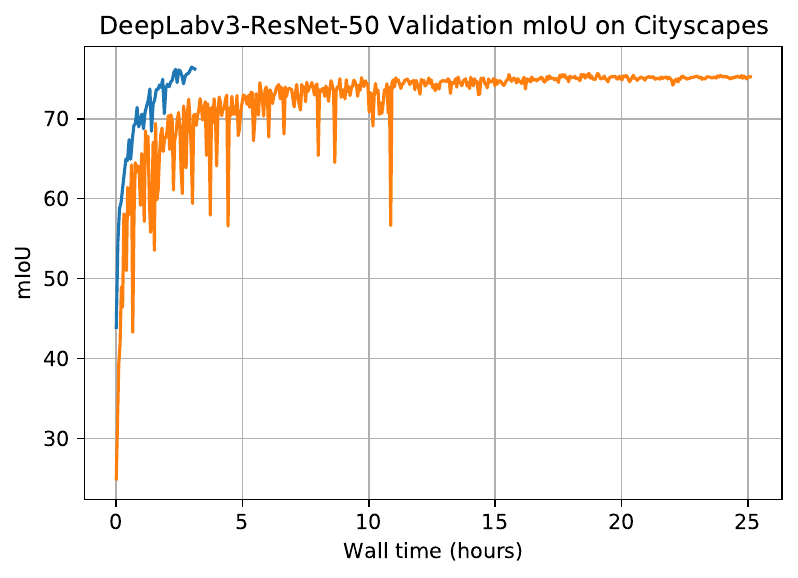}}
\caption[cityscapes supp]{(a) Training on the Cityscapes dataset from scratch for 200 epochs. (b) Fine-tuning using a backbone pretrained on Imagenet for 50 epochs with ND++ and 500 epochs with SyncBN. While each epoch takes longer with ND++ layers, the training convergence is still markedly faster in terms of wall time. Note that we use a learning rate of 0.1 for the ND++ model here.}\label{fig:cityscapes_supp}
\end{figure}
 
\begin{figure}[hbt!]
\includegraphics[width=.99\linewidth]{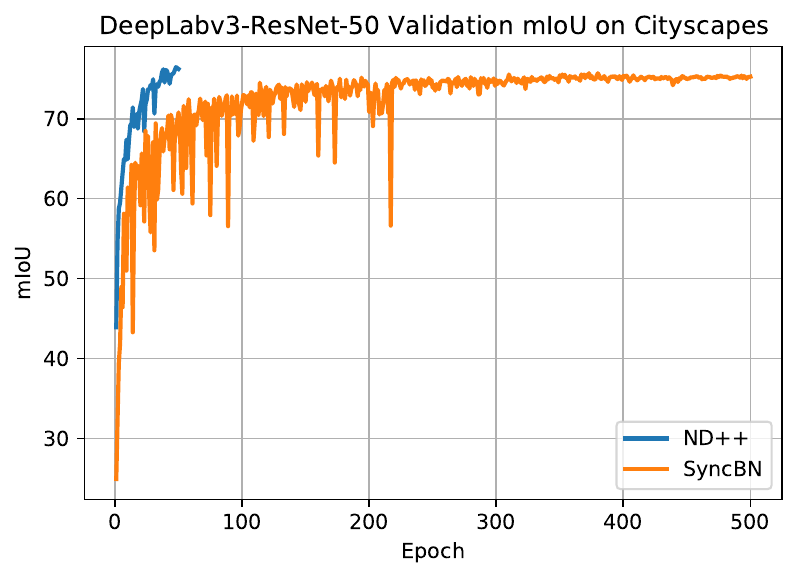}
\caption[front page]{mIoU curves of DeepLabv3 with a ResNet-50 backbone trained on the Cityscapes dataset using ND++ and SyncBN. Here, we show that ND++ takes less than 50 epochs to achieve the same accuracy as SyncBN achieves in 500 epochs.}
\label{fig:one-tenth}
\end{figure}

We see accelerated training and improved final performance on the validation set both when training from scratch and when fine-tuning with a pretrained backbone (Supp. Fig.~\ref{fig:cityscapes_supp} (a)). While the model equipped with ND++ layers takes approximately 30\% longer per epoch to train, the accelerated training convergence more than makes up for the increased time per epoch (Fig.~\ref{fig:cityscapes_supp} (b)).

Similar to object detection, we find cross-GPU synchronization markedly improves performance (from $69.82\%$ to $72.69\%$) for semantic segmentation (Supp. Table~\ref{tab:SemSegAblation}). However, unlike for object detection, enforcing scale invariance has very little effect. Given this fact, we report results without enforcing scale invariance in Table~\ref{tab:Cityscapes} and Fig.~\ref{fig:frontpage}.

\subsection{Training Transformer Models for Language Tasks}

\begin{figure}[hbt!]
\centering
\subfigure[]{\includegraphics[width=.49\columnwidth]{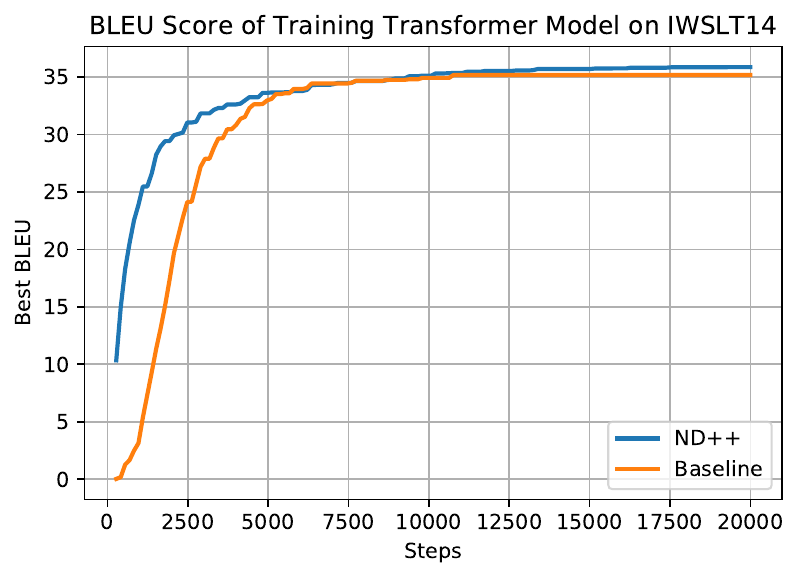}}
\subfigure[]{\includegraphics[width=.49\columnwidth]{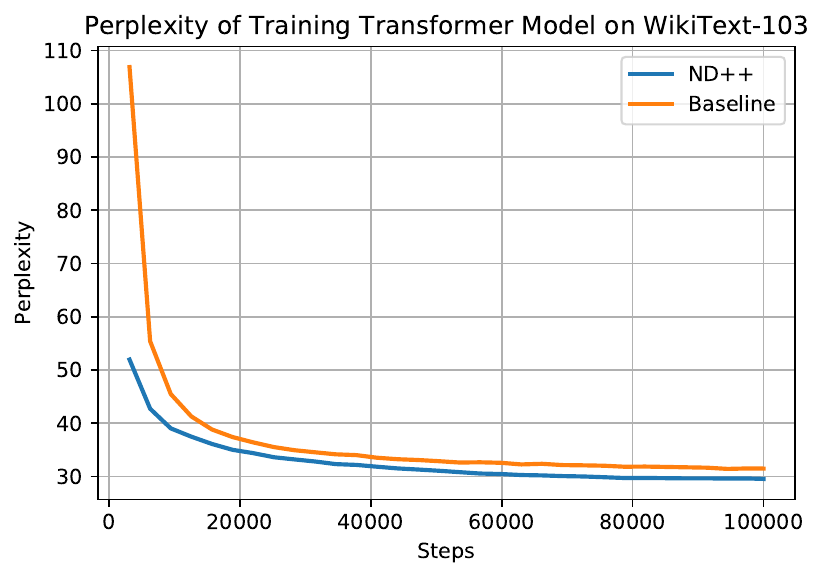}}
\caption[NLP]{Training conducted using ND++ with a cosine learning rate decay versus the baseline with warmup and inverse square root decay. (a) BLEU score comparison on IWSLT14. (b) Perplexity comparison on  WikiText-103.}\label{fig:cityscapes_supp}
\end{figure}

We also find promising results when training two Transformer models~\cite{vaswani2017attention} for language tasks using the code  from Fairseq
. We remove the Layer Normalization layers in the baseline Transformer model and insert our modifications in the linear transform layers. The original transformer models require warmup iterations for successful training. After modification with ND++, we can start with a large learning rate and remove the warmup iterations. For the German to English translation task on the IWSLT14 dataset, the BLEU score of ND++ after 20k iterations is 35.9 vs 35.2 of the baseline. For the language model training on WikiText-103, the Perplexity is reduced from 31.48 of the baseline model to 29.54. For these experiments the max token size is 4096 per GPU and we train on 8 Nvidia A100 GPUs. The baseline models are trained with the inverse square root learning decay after 4k warmup iterations. After modifying with ND++, we notice using a simple cosine learning rate decay as in vision tasks significantly accelerates convergence.

 \section{Relation to Other Decorrelation Methods}
\label{sec:relation}

\begin{figure}[hbt!]
\centering
\includegraphics[width=.8\columnwidth]{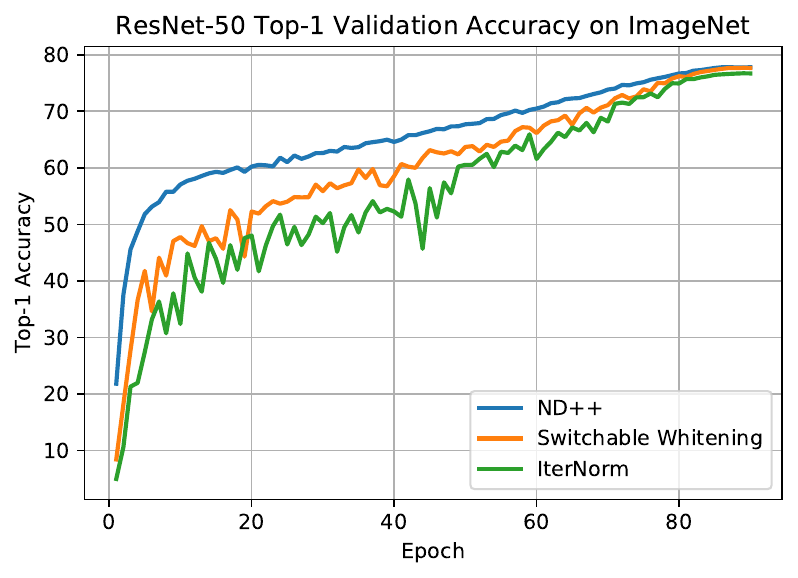}
\caption[extra experiments]{Top-1 validation accuracy curves of ResNet-50 when trained on ImageNet with ND++, Switchable Whitening(SW), IterNorm using batch size of $2048$.}
\label{fig:whitening}
\vspace{-1.5em}
\end{figure}

\begin{table}[]
\resizebox{0.45\textwidth}{!}{%
\begin{tabular}{cc|ccc}
\multicolumn{2}{c|}{CIFAR-100}                         & \multicolumn{3}{c}{ImageNet}                                                     \\
\multicolumn{2}{c|}{ResNet-110}                        & \multicolumn{3}{c}{ResNet-50}                                                    \\ \hline
SW (160 epochs)                & ND++ (80 epochs)                & IterNorm                  & SW                        & ND++                      \\
\multicolumn{1}{r}{73.52} & \multicolumn{1}{r|}{\textbf{74.74}} & \multicolumn{1}{r}{76.69} & \multicolumn{1}{r}{77.66} & \multicolumn{1}{r}{\textbf{77.95}}
\end{tabular}%
}
\caption[rebuttal]{Extra results on CIFAR-100 and ImageNet.}
\label{tab:whitening}
\vspace{-1em}
\end{table}

We highlight a few differences between our work and recent whitening techniques~\cite{huang2018decorrelated,huang2019iterative,pan2019switchable}. Our deconvolution is a normalization in the frequency domain. This is the optimal feature transform for convolution operation, while recent works are developed for non-convolutional operations. We follow the simple yet important mathematical justification to insert it before the weights, removing the redundancy and avoiding potential problems related to extra parameters as in BatchNorm. Moreover, we avoid the computational cost of explicit decorrelation by fusing the decorrelation matrix with the weight matrix in the linear transform. Without this, per-sample whitening could incur high computational cost. Due to this limitation, most recent works~\cite{huang2018decorrelated,huang2019iterative,pan2019switchable} have restricted the use of whitening to only a few layers of the network. Additionally, the coupled inverse newton iteration method we adopt, along with the Denman-Beavers iteration in~\cite{DBLP:journals/corr/abs-1809-08625}, are stable to finite arithmetic while the vanilla Newton-Schulz method diverges under certain conditions~\cite{Ye2020Network}(Fig. 7).  

We compare our results with IterNorm and Switchable Whitening(SW) in various settings according to their code availability. In all settings ND++ achieves faster and better convergence than IterNorm and SW(Fig.~\ref{fig:whitening}). Interestingly, we found our simple design performs slightly better than the ensemble design of SW that incorporates multiple normalization/whitening techniques while using less computational space. This suggests that the ensemble design has potential for further improvement by including ND++ as a module (Table.~\ref{tab:whitening}). When training the ResNet-50 with a learning rate of 1.0 and no momentum on the ImageNet, ND++ achieves a top accuracy of 77.66 versus 76.85 for SW and 76.57 for IterNorm. We tried to compare the results of training ResNet-101 but SW ran out of GPU RAM. Nevertheless, our number of 79.4 with 90 epochs of training is higher than the 78.6 reported in their paper with 100 epochs of training.

\end{document}